%% file: main.tex
\documentclass[runningheads]{llncs}

\usepackage[numbers,sort&compress,sectionbib]{natbib} 
\AtBeginDocument{}
\usepackage{graphicx}
\usepackage[utf8]{inputenc}
\usepackage[T1]{fontenc}
\usepackage{xcolor}

\begin{document}
\title{Building Metadata Inference Using a Transducer Based Language Model}


\author{David Waterworth\inst{1}\orcidID{0000-0002-5262-3803} \and
Subbu Sethuvenkatraman\inst{2}\orcidID{0000-0001-7197-2307} \and
Quan Z. Sheng\inst{1}\orcidID{0000-0002-3326-4147}}
\authorrunning{D. Waterworth et al.}
%
\institute{School of Computing, Macquarie University, Sydney, NSW, Australia
\email{\{david.waterworth,michael.sheng\}@mq.edu.au}\\
\and Energy Business Unit, Commonwealth Scientific and Industrial Research Organisation (CSIRO), Newcastle, NSW, Australia\\
\email{subbu.sethuvenkatraman@csiro.au}}

\maketitle              
\begin{abstract}
Solving the challenges of automatic machine translation of Building Automation System text metadata is a crucial first step in efficiently deploying smart building applications. The vocabulary used to describe building metadata appears small compared to general natural languages, but each term has multiple commonly used abbreviations. Conventional machine learning techniques are inefficient since they need to learn many different forms for the same word, and large amounts of data must be used to train these models. It is also difficult to apply standard techniques such as tokenisation since this commonly results in multiple output tags being associated with a single input token, something traditional sequence labelling models do not allow. Finite State Transducers can model sequence-to-sequence tasks where the input and output sequences are different lengths, and they can be combined with language models to ensure a valid output sequence is generated. We perform a preliminary analysis into the use of transducer-based language models to parse and normalise building point metadata.

\end{abstract}

\setcounter{footnote}{0} %
\input{body}

\subsubsection{Acknowledgements.} 

The work of David Waterworth is funded by the RoZetta Institute, the CSIRO and CIM Pty Ltd. Quan Z. Sheng's work has been partially supported by an Australian Research Council (ARC) Linkage Infrastructure, Equipment and Facilities Projects ARC LE180100158 and LE220100078.

\bibliographystyle{splncs04nat}
\bibliography{references}
\end{document}

%% file: body.tex
\section{Introduction}


The International Energy Agency~(IEA) has estimated that buildings were responsible for 28\% of global emissions in 2019~\cite{IEA:2020}. Commercial buildings are a significant energy consumer, responsible for more than 7\% of global final energy consumption and up to 18\% in some industrialised economies~\cite{PrezLombard:2008}. A typical commercial building has been shown to waste 30\% of its energy consumption~\cite{Pritoni:2021}. Incorporating digital technologies in buildings can reduce energy consumption, increase operational efficiency~\cite{Hong:2020}, and improve occupant comfort and productivity~\cite{Zhang:2017}.

A Building Automation System (BAS) consists of `points' representing `sensors', `actuators', `settings' and `setpoints'. These points are described by short text fields containing unstructured metadata that may provide information such as sensor type and location and what equipment it is associated with. Novel artificial intelligence (AI) and machine learning (ML) solutions for managing buildings are being developed, such as occupancy detection~\cite{Acquaah:2020}, fault detection and diagnosis~\cite{Tun:2021}, and grid support services~\cite{Wang:2021}. However, deployment of these applications and algorithms is hamstrung because the BAS point metadata was not designed to support these tasks~\cite{Bhattacharya:2015}. Due to device memory limitations and to reduce input keystrokes, point text metadata is often heavily abbreviated, and the abbreviated forms can vary significantly from building to building. Mapping to one of many machine-readable schemas, such as Brick~\cite{Balaji:2016} or Haystack~\cite{Praire:2016}, is required, and this is usually performed manually.

Machine learning approaches to automate metadata mapping show promise. However, they require large amounts of data from many buildings in order to learn the multitude of ways engineers can express the same concept, and they're prone to making unrecoverable mistakes~\cite{Gorman:2021}. In this paper, we use a hybrid approach of slot rules defined by finite-state transducers (FSTs) combined with a statistical language model. This approach allows us to more easily include expert knowledge in our model whilst retaining the power of statistical techniques to learn from data.


\section{Prior Work}

Most studies into the BAS metadata mapping focus on reducing the cost of manually labelling each building independently. The goal is to select the smallest subset of points $\mathcal{D}_l$ from the initially unlabelled dataset $\mathcal{D}_u$ that maximise the accuracy a model $y=f(x)$ where $(x,y) \in \mathcal{D}_l$. \emph{Active learning}~\cite{Settles:2009} is used to iteratively generate a set of \emph{representative samples} $\mathcal{D}_u$ which an expert then labels.

One of the earliest attempts to automate metadata mapping used a rules-based approach~\cite{Bhattacharya:2014a, Bhattacharya:2015} using synthesis by example~\cite{Gulwani:2012}. Later approaches focus on statistical machine learning. Sequence labelling is used to split the text into non-overlapping chunks and apply class labels to each chunk~\cite{Koh:2018a,Lin:2019,Jiao:2020}. This approach combined with active learning works well since the same abbreviations are used consistently between points from the same building. These approaches all train a \emph{local model}, a model that is trained on a single building only. There have been some attempts to train \emph{global models} on multiple buildings~\cite{Ma:2020,Waterworth:2021}. Most of these approaches have only been validated on small datasets consisting of 3-5 buildings so it is unclear how they scale to larger datasets.

Finite-state transducers (FSTs) have been used with language models to expand abbreviations~\cite{Sproat:2016, Gorman:2021}. FSTs are also used in speech applications because they can be constructed to ensure that the output sequence is valid~\cite{Mohri:2002}.

In this work, we perform a preliminary investigation into the use of FSTs combined with a statistical language model to extract semantic information from BAS text metadata. To the best of our knowledge, we are the first to use a hybrid rules-based and statistical approach to this task. 

\section{The Metadata Mapping Task}\label{sec:mapping}

There are two aspects of the metadata mapping task. The first is assigning \emph{tags} to sequences of abbreviations. A tag is a single piece of semantic information such as \texttt{air}, \texttt{temperature} and \texttt{sensor}. A collection of tags is known as a \emph{tagset}. The second task is identifying text chunks corresponding to physical entities such as equipment and locations. The challenges with this approach are two-fold. First, it is common for entities to nest -- `AHU-L01-02' is an entity of type \texttt{Air\_Handling\_Unit} containing a nested \texttt{Level} entity. Second, a single input token may map to multiple output tags -- `SAF' should be labelled with three tags, \texttt{supply}, \texttt{air} and \texttt{fan}. Prior work uses character-based models and multiple inference steps. This introduces inefficiencies that we address by using a Finite State Transducer slot model with a Language Model.

Two idiosyncratic practices define BAS point text descriptions and make it difficult for automated processes to extract metadata. The first is the heavy use of abbreviations. In a portfolio of over 200 commercial buildings, we observed the following:

\begin{itemize}
    \item \textbf{Initialism:} A group of letters, each pronounced separately. 
    \item \textbf{Truncation:} This type of abbreviation consists only of the first $n$ characters of a word. 
    \item \textbf{Vowel Deletion:} All vowels are removed, this may be combined with truncation. 
    \item \textbf{Contraction:} Only the first and last letter is retained.
    \item \textbf{Syllable Division:} Words are split into syllables then one or more of the previous rules are applied to each syllable. 
    \item \textbf{Phonetic:} Similar to syllable division using phonemes. 
\end{itemize}

The second is deletion or substitution of whitespace with special characters, we have observed the following practices:

\begin{itemize}
    \item \textbf{camelCase:} The first letter of each word other than the first word is upper case. All whitespace is deleted.
    \item \textbf{PascalCase:} The first letter of each word is upper case. All whitespace is deleted.
    \item \textbf{snake\_case:} Whitespace is replaced with an underscore (\_).
    \item \textbf{kebab-case:} Whitespace is replaced with a dash (-).
    \item \textbf{UPPERCASE:} All whitespace is deleted and all characters converted to upper case.
\end{itemize}

These practices make it very difficult to use rules-based methods such as regular expressions to extract information. For example, `AHU-01A' is a valid entity by itself, but if it appears in the text `AHU-01AhrsMd' then the correct prediction is `AHU-01' should be labelled as \textsf{Air\_Handling\_Unit} and `AhrsMd' as the tagset containing tags \textsf{after}, \textsf{hours} and \textsf{mode}. 

\section{Finite State Transducer and Language Model}

\begin{figure}[ht!]
    \centering
    \includegraphics[width=0.8\textwidth]{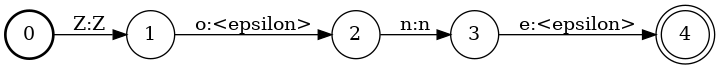}
    \caption{An example of a finite-state transducer that abbreviates the string `Zone' to `Zn' by deleting vowels. The $\epsilon$ symbol represents no output for the deleted vowels `o' and `e'.}.
    \label{fig:fst-zone}
\end{figure}

A finite-state acceptor (FSA) is a state machine with no outputs. It consists of a finite alphabet of input symbols, a finite set of states, a starting state, one or more final (accepting) states and a transition function. An FSA is said to accept a sequence of input symbols if there exists a sequence of valid transitions from the start state to a final state. 

Finite-state transducers (FSTs) are generalisations of FSAs that represent the relationship between sets of strings. Each transition includes both an input and output symbol. In addition, the special character $\epsilon$ is added to represent no input or no output, allowing relationships between different length strings to be represented. For example, the FST shown in Figure~\ref{fig:fst-zone} transduces `Zone' to `Zn'.


Weighted finite-state transducers (WFSTs) are FSTs where the transitions have an associated weight in addition to the input and output symbols. The weights can be log probabilities, allowing different paths from the starting state to each final state to be scored based on maximum likelihood by weight summation. We use this property to generate a lattice that contains multiple potential output sequences and then use a language model to select the \emph{most likely} sequence. 


\subsection{Finite State Transducer Slots}

\begin{figure}[htbp]
    \centering
    \includegraphics[width=0.8\textwidth]{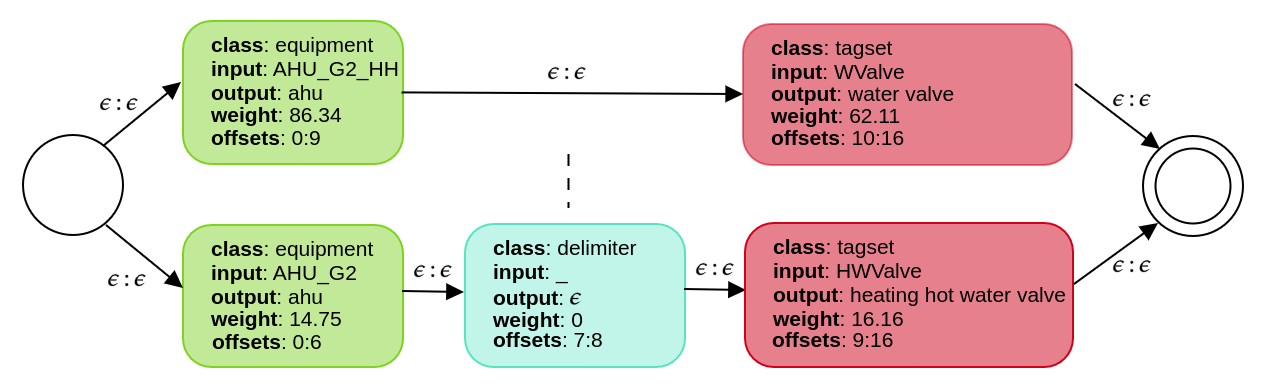}
    \caption{An example of a slot lattice generated from the input text \texttt{AHU\_G2\_HHWValve}. We have only shown two paths in order to preserve space. Each slot comprises a weighted finite-state transducer from the input to the output, with an assigned weight. The FSTs from each slot are combined into a single FST. This allows us to efficiently extract the path with maximum likelihood.}
    \label{fig:slots}
\end{figure}

A \emph{slot} is a class representing matched text. Each slot consists of a weighted finite-state transducer. The transducer outputs one or more tags, or the empty string ($\epsilon$). Each slot has a weight, and slots form a directed graph or lattice. There is an arc between slots if the text they match is adjacent in the original string. The slot weights are either fixed, or dynamically assigned by a language model depending on the slot class. \\

\noindent Construction of the lattice proceeds as follows:

\begin{enumerate}
\item All substrings of the input string are enumerated.
\item Each substring is tested against each slot class. If a class accepts a substring, a slot is created for every output string generated from the input.
\item The FSTs of adjacent slots are joined.
\item A start and final node are added.
\item The weights are updated using the language model for dynamic slots.
\item The shortest path is calculated.
\item The slots from the shortest path are extracted.
\end{enumerate}

\noindent We used the following slots:


\subsubsection{delimiter} The delimiter slot matches one of \texttt{\textcolor{blue}[ -\_.]} and outputs the empty string symbol (\texttt{<epsilon>}).

\subsubsection{equipment} The \emph{equipment} slot matches \texttt{<}\textcolor{blue}{\texttt{type}}\texttt{>} \texttt{<}\textcolor{blue}{\texttt{delimiter}}\texttt{>}? \texttt{<}\textcolor{blue}{\texttt{identifier}}\texttt{>} (\texttt{<}\textcolor{blue}{\texttt{delimiter}}\texttt{>} \texttt{<}\textcolor{blue}{\texttt{identifier}}\texttt{>})* and outputs the equipment type tag.

\subsubsection{tagset}

The \emph{tagset} slot matches point descriptions, both abbreviated and unabbreviated. The matched text can also include delimiter characters, and the output is a sequence of expanded tags; for example inputting \texttt{SAF} results in the output \texttt{supply air fan}. Multiple tagset slots may be generated for the same input as abbreviations are often ambiguous.

\subsubsection{unknown}

The final rule matches any alphanumeric character and outputs \texttt{<epsilon>}.

\subsection{Language Model}

A language model is a probabilistic model that enables us to score a tagset based on the likelihood of the sequence of symbols. The \texttt{equipment} and \texttt{tagset} slots each have an associated language model. The \texttt{tagset} language model is trained on HVAC phrases such as \texttt{zone temperature setpoint} and the \texttt{equipment} language model is a character model trained on equipment names such as \texttt{AHU-01}. In both cases, a count-based language model of order=2 was used, and perplexity used to compute the slot weights. The weight of the \texttt{delimiter} slot is fixed at zero, and the weight of the \texttt{unknown} slot is a large constant value.





\section{Experiment}

In order to evaluate the feasibility of using FSTs to model BAS point metadata, we manually labelled 3,000 randomly selected points from 10 commercial buildings. Due to resource limitations, we choose only to label air handling units (including fan coil units and packaged units). As previously described, we used this data to construct an FST parser and language model. Finally, we tested the model on air handling unit points from five buildings that were not part of the dataset used to build the model. The buildings were diverse; some used abbreviations such as `ZnT' for `zone temperature' and others longer descriptions such as `RmTemp\_Setpt' for ‘room temperature setpoint’. The structure of points from the different buildings was also quite different, ranging from `AHU\_G\_1\_Mode' to `B550-NAE19/N2-2.DX-103.B550\_AHU\_02-03 CV-C'.

Accuracy ranged from 88.6\% to 93.1\%. There were two sources of errors. The first was out-of-vocabulary. Some points such as `minimum outside air damper' (MOAD) were not included in the training set. A secondary issue was due to equipment such as `RAF\_19\_2' being labelled as a tagset (return air fan) rather than as equipment. This is because our dataset contains return air fan points labelled as air handling units (since they are a component of an air handling unit) and we did not include fans when training the equipment language model. We observed that it was easy to rectify these errors by adding the missing terms and regenerating the model, a process that can be done online.

\section{Discussion and Conclusions}

In this study, we have demonstrated that it is possible to use a hybrid of rules with a language model to train a global model to extract semantic information from BAS text metadata. 

Given the open nature of the problem at hand, training a single model that can correctly classify any BAS (Building Automation System) point is very difficult. This is why most of the prior work focuses on identifying the smallest set of points that require manual annotation through the use of active learning.

A promising alternative is continual learning, where a single model trained on multiple buildings is continuously updated with new information. To the best of our knowledge, this approach has yet to been studied using BAS metadata. To do so effectively would require a model that can be easily updated, ideally using only one or a few new examples.

A finite-state transducer abbreviation model shows promise in this area. Most of the errors we identified could be easily corrected by updating the abbreviation models vocabulary with new abbreviated words and phrases as they are encountered. Despite these errors our model preformed very well when trained on a relatively small dataset. Similar to active learning, when the model encounters out-of-vocabulary tokens, it could prompt the user to provide an expansion. In most cases, this would require the user to enter a short phrase. Since the abbreviation model is not gradient-based, it can be rapidly retrained.
